\pdfoutput=1
\documentclass[conference]{IEEEtran}
\usepackage{cite}
\usepackage{graphicx}
\usepackage{booktabs}
\usepackage{amsmath}
\usepackage{amssymb}
\usepackage{array}
\usepackage{url}
\usepackage[caption=false,font=footnotesize]{subfig}
\usepackage[dvipsnames]{xcolor}
\usepackage{tikz}
\usetikzlibrary{arrows.meta,positioning,shapes.geometric}
\graphicspath{{plots/}}

\title{AegisUI: Behavioral Anomaly Detection for Structured User Interface Protocols in AI Agent Systems}

\author{
\IEEEauthorblockN{Mohd Safwan Uddin$^{1}$, Saba Hajira$^{2}$}
\IEEEauthorblockA{$^{1}$Department of Computer Science, $^{2}$Department of Information Technology\\
Muffakham Jah College of Engineering and Technology, Hyderabad, India\\
Email: $^{1}$safwanuddin405@gmail.com, $^{2}$sabahajira422@gmail.com}
}

\begin{document}
\maketitle

\begin{abstract}
AI agents that build user interfaces on the fly assembling buttons, forms, and data displays from structured protocol payloads are becoming common in production systems. The trouble is that a payload can pass every schema check and still trick a user: a button might say ``View invoice'' while its hidden action wipes an account, or a display widget might quietly bind to an internal salary field. Current defenses stop at syntax; they were never built to catch this kind of behavioral mismatch.

We built AegisUI to study exactly this gap. The framework generates structured UI payloads, injects realistic attacks into them, extracts numeric features, and benchmarks anomaly detectors end-to-end. We produced 4{,}000 labeled payloads (3{,}000 benign, 1{,}000 malicious) spanning five application domains and five attack families: phishing interfaces, data leakage, layout abuse, manipulative UI, and workflow anomalies.

From each payload we extracted 18 features covering structural, semantic, binding, and session dimensions, then compared three detectors: Isolation Forest (unsupervised), a benign-trained autoencoder (semi-supervised), and Random Forest (supervised). On a stratified 80/20 split, Random Forest scored best overall (accuracy 0.931, precision 0.980, recall 0.740, F1 0.843, ROC-AUC 0.952). The autoencoder came second (F1 0.762, ROC-AUC 0.863) and needs no malicious labels at training time, which matters when deploying a new system that lacks attack history. Per-attack-type analysis showed that layout abuse is easiest to catch while manipulative UI payloads are hardest. All code, data, and configs ship with the paper for full reproducibility.
\end{abstract}

\section{Introduction}
A few years ago, an AI agent was just a chatbot. You typed a question, it returned text. Now agents plan multi-step workflows, call external APIs, and increasingly generate the user interface itself \cite{react2022,toolformer2023,autogpt2023}. Instead of hand-coded screens, the agent emits a structured payload describing what to render: which buttons, which forms, which data sources, what layout. The renderer on the client side follows the instructions.

This pattern shows up in several emerging protocol designs, including Agent-to-User Interface (A2UI)-style specifications where the agent produces a declarative description and a generic renderer interprets it \cite{a2ui_spec}. It is a clean separation of concerns and it works well until someone tampers with the payload.

The core problem is straightforward. A protocol payload can satisfy every structural constraint (valid JSON, correct field types, required keys present) and still be dangerous. We kept running into three scenarios while designing the framework:

\begin{enumerate}
    \item A booking assistant generates a payment form. An attacker injects extra fields such as ``corporate email verification'' and ``password'' into the same container. The payload remains schema-valid. The user sees what looks like a normal checkout page.
    \item An analytics dashboard binds a display widget to \texttt{internal.salary\_band} instead of the intended aggregated metric. The layout looks identical. No schema rule catches a wrong binding target.
    \item A workflow approval screen shows an ``Approve'' button whose hidden action is \texttt{delete\_account}. The label and the behavior are completely disconnected, but nothing in the type system flags it.
\end{enumerate}

We looked for existing tools or datasets to study these threats and found very little. Network intrusion detection has NSL-KDD and CICIDS2017 \cite{nslkdd,cicids2017}. Phishing detection has web-page datasets \cite{garera2007framework,chiew2019web}. But for structured agent-generated UI protocols? Nothing. No benchmark, no labeled dataset, no end-to-end experimental pipeline.

So we built one. AegisUI covers the full loop: payload generation with domain-aware blueprints, adversarial mutation across five attack families, schema and logical validation, feature extraction into an 18-dimensional numeric vector, and comparative evaluation of three detection models. Everything is seed-controlled and deterministic: run it twice, get the same dataset.

Our contributions:
\begin{itemize}
    \item A payload generation system that produces realistic UI protocol objects across five domains (booking, e-commerce, analytics, forms, workflow approval), with targeted attack injection via mutation of benign seeds.
    \item A labeled dataset of 4{,}000 payloads with traceable attack metadata. Every malicious sample records exactly which components were injected or modified and why.
    \item An 18-feature extraction pipeline organized into structural, semantic, binding, and session groups.
    \item A three-model comparison (Isolation Forest, autoencoder, Random Forest) with per-attack breakdowns and a feature-group ablation study.
\end{itemize}

We want to be upfront about scope. The dataset is synthetic. We have not tested on production traffic from a real agent system. What we have is a controlled environment where the ground truth is exact, the experiments are reproducible, and the baseline numbers are credible enough to build on.

\section{Problem Statement}
Traditional web UIs are static. A developer writes HTML, reviews it, ships it. Security checks happen at code review, at deployment, at the API boundary. When the UI itself is generated at runtime by an agent, that chain breaks. The payload carries the logic: component hierarchy, data bindings, action definitions, and if the payload is compromised, the rendered interface becomes the attack vector \cite{owasp,nistairmf}.

Schema validation does not solve this. A schema says: ``the \texttt{components} field must be an array of objects, each with a \texttt{component\_type} string and a \texttt{label\_text} string.'' It cannot say: ``a button labeled `Continue safely' should probably not trigger \texttt{delete\_account}.'' That second check requires reasoning about behavioral consistency, not type correctness.

We frame the detection task as follows. Let $\mathcal{P}$ be the set of schema-valid payloads. We define a feature map $\Phi: \mathcal{P} \rightarrow \mathbb{R}^{d}$ converting each payload into a numeric vector capturing its structural and semantic properties. A scoring function $S: \mathbb{R}^{d} \rightarrow \mathbb{R}$ assigns a risk value, and a threshold $\tau$ produces a binary decision:
\begin{equation}
\hat{y}=
\begin{cases}
1 \text{ (malicious)}, & \text{if } S(\Phi(p)) \ge \tau\\
0 \text{ (benign)}, & \text{otherwise.}
\end{cases}
\end{equation}

For the supervised case (Random Forest), we minimize empirical risk with cross-entropy loss over labeled pairs $\{(x_i, y_i)\}_{i=1}^{N}$:
\begin{equation}
\min_{\theta}\ \frac{1}{N}\sum_{i=1}^{N}\mathcal{L}\bigl(f_{\theta}(x_i),\, y_i\bigr)+\lambda\,\Omega(\theta).
\end{equation}
For the autoencoder, $S$ is the mean squared reconstruction error $\|x - \hat{x}\|_2^2$ trained on benign data alone, and $\tau$ is set at the 95th percentile of benign training errors. For Isolation Forest, $S$ is the negated isolation score (shorter average path length signals stronger anomaly).

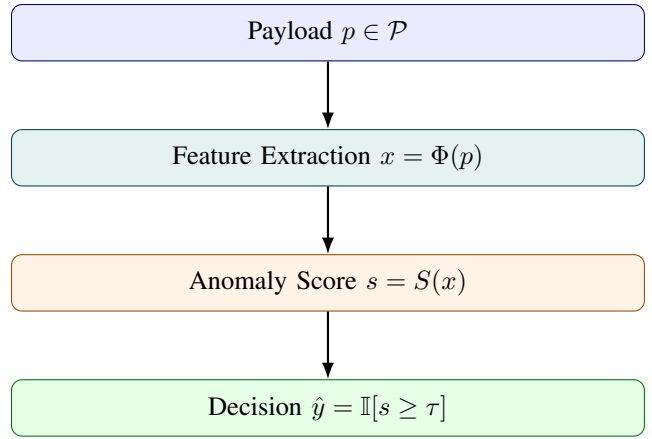
\begin{figure}[t]
\centering
\begin{tikzpicture}[
node distance=0.9cm and 0.6cm,
blk/.style={draw, rounded corners, align=center, minimum width=0.95\linewidth, minimum height=0.75cm},
arr/.style={-{Latex[length=2.2mm]}, thick}
]
\node[blk, fill=blue!8, draw=blue!55!black] (p) {Payload $p \in \mathcal{P}$};
\node[blk, fill=teal!10, draw=teal!55!black, below=of p] (phi) {Feature Extraction $x=\Phi(p)$};
\node[blk, fill=orange!10, draw=orange!60!black, below=of phi] (s) {Anomaly Score $s=S(x)$};
\node[blk, fill=green!10, draw=green!45!black, below=of s] (d) {Decision $\hat{y}=\mathbb{I}[s \ge \tau]$};
\draw[arr] (p) -- (phi);
\draw[arr] (phi) -- (s);
\draw[arr] (s) -- (d);
\end{tikzpicture}
\caption{Detection pipeline: payload to feature vector to anomaly score to binary decision.}
\label{fig:math-pipeline}
\end{figure}

\section{Related Work}
\textbf{Anomaly detection in security.}
Chandola et al.\ \cite{chandola2009anomaly} survey anomaly detection broadly. In intrusion detection, Liao et al.\ \cite{liao2013intrusion} catalogue the shift from signature matching to statistical and ML-based methods. Sommer and Paxson \cite{somer2010outside} make a sobering point: anomaly detectors that look great on benchmarks often falter in deployment because benign traffic is messy and adversaries adapt. We took that warning seriously when interpreting our own results. It is one reason we test three model families rather than reporting only the best one, and it is why we are cautious about generalizing from synthetic data.

\textbf{Isolation Forest and autoencoders.}
Liu et al.\ \cite{liu2008isolation} introduced Isolation Forest, which isolates outliers using random partitioning of the feature space; points that separate quickly are likely anomalies. It needs no labels, which is its main selling point. Sakurada and Yairi \cite{sakurada2014anomaly} demonstrated that autoencoders detect anomalies through reconstruction error, and Chalapathy and Chawla \cite{chalapathy2019deep} survey the broader deep anomaly detection landscape. We apply both methods; the autoencoder in our case trains exclusively on benign payloads, so it should reconstruct normal patterns well and produce high error on attack payloads.

\textbf{Supervised baselines.}
Random Forest \cite{breiman2001random} remains a reliable workhorse in applied ML. Zhang et al.\ \cite{idsml} showed it performs well on network intrusion data even with mixed feature types. We use it as an upper baseline: when labels are available, what accuracy can we expect?

\textbf{Interface-level threats.}
Dhamija et al.\ \cite{phishinghuman} studied why phishing works from a human-factors perspective: users rely on surface cues and miss subtle inconsistencies. Garera et al.\ \cite{garera2007framework} built classifiers for phishing URLs, and Chiew et al.\ \cite{chiew2019web} used hybrid feature selection for phishing page detection. These are the closest relatives to our work, but they target rendered web pages, not the structured protocol payloads that an agent sends before rendering begins.

\textbf{Agent systems and generated interfaces.}
The ReAct framework \cite{react2022} showed that interleaving reasoning and tool actions lets agents handle complex tasks. Toolformer \cite{toolformer2023} taught language models to invoke tools autonomously. Auto-GPT \cite{autogpt2023} demonstrated fully autonomous task loops. As these systems mature, the UIs they produce become a trust boundary, and we could not find prior work treating agent-generated protocol payloads as a security object with a proper detection benchmark.

\textbf{Gap.}
Public benchmarks like NSL-KDD \cite{nslkdd} and CICIDS2017 \cite{cicids2017} cover network traffic. Phishing datasets cover rendered pages. Neither captures the structure of an agent-to-renderer protocol: the component trees, action bindings, layout graphs, and session metadata that make up our feature space. AegisUI fills that gap.

\section{Threat Model}
The adversary operates at the protocol layer. They do not need to break encryption, steal credentials, or compromise the host. They need to get a crafted payload into the rendering pipeline, through prompt injection, a compromised agent plugin, a man-in-the-middle on an internal API, or a poisoned tool response.

Their payloads pass schema validation. Required fields exist, types match, IDs are consistent. What differs is behavior: bindings point to sensitive internal data, action labels mismatch actual operations, or layout structures grow abnormally deep and large.

What is at stake:
\begin{itemize}
    \item \textbf{User credentials and payment data}, targeted by phishing-style field injection.
    \item \textbf{Internal records}, exposed by binding display widgets to privileged sources like \texttt{internal.user\_ssn} or \texttt{internal.auth\_secret}.
    \item \textbf{Workflow integrity}, broken by reordering actions so approvals fire before required inputs are collected or disguising destructive operations behind benign labels.
\end{itemize}

The defender sees the payload before it hits the renderer. They can inspect components, the layout graph, bindings, timestamps, and session IDs. They cannot read the user's mind or access runtime interaction traces. This is a pre-render screening problem.

\begin{figure}[t]
\centering
\begin{tikzpicture}[
node distance=0.8cm and 0.4cm,
blk/.style={draw, rounded corners, align=center, minimum width=0.95\linewidth, minimum height=0.72cm},
arr/.style={-{Latex[length=2mm]}, thick}
]
\node[blk, fill=purple!10, draw=purple!60!black] (agent) {Agent / Payload Producer};
\node[blk, fill=red!10, draw=red!55!black, below=of agent] (boundary) {Untrusted Protocol Channel};
\node[blk, fill=cyan!12, draw=cyan!55!black, below=of boundary] (def) {AegisUI Detector\\(schema check + behavioral scoring)};
\node[blk, fill=green!10, draw=green!45!black, below=of def] (user) {Renderer + End User};
\draw[arr] (agent) -- (boundary);
\draw[arr] (boundary) -- (def);
\draw[arr] (def) -- (user);
\end{tikzpicture}
\caption{Trust boundaries. The detector sits between the protocol channel and the renderer.}
\label{fig:trust-boundary}
\end{figure}
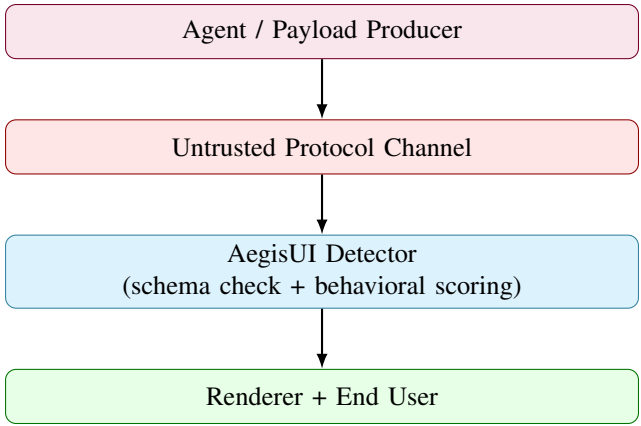

We scope this work to pre-render detection only. Post-interaction forensics, endpoint telemetry, and user behavior modeling are separate problems outside our current focus.

\section{AegisUI Framework}
The pipeline has four stages: generate, validate, extract, detect. Each stage reads the output of the previous one and writes its own. A single YAML config and a global seed (1337) control the entire run, so re-executing produces identical artifacts.

\begin{figure*}[t]
\centering
\resizebox{\textwidth}{!}{
\begin{tikzpicture}[
    node distance=1.8cm and 1.2cm,
    box/.style={draw, rounded corners, align=center, minimum width=3.1cm, minimum height=1.1cm},
    io/.style={draw, rounded corners, fill=gray!10, align=center, minimum width=3.1cm, minimum height=1.1cm},
    arr/.style={-{Latex[length=2.6mm]}, thick}
]
\node[io, fill=blue!8, draw=blue!55!black] (cfg) {Config + Seed\\dataset\_config.yaml};
\node[box, fill=teal!10, draw=teal!55!black, right=of cfg] (gen) {Payload Generator\\(benign + attack)};
\node[box, fill=orange!12, draw=orange!55!black, right=of gen] (val) {Schema + Logical\\Validation};
\node[box, fill=purple!10, draw=purple!55!black, right=of val] (feat) {Feature Extraction\\(18 numeric features)};
\node[io, fill=green!12, draw=green!45!black, right=of feat] (det) {Detection Models\\IF / AE / RF};
\node[io, fill=cyan!10, draw=cyan!50!black, below=of gen] (dset) {Dataset Store\\benign/malicious JSON};
\node[io, fill=yellow!14, draw=yellow!55!black, below=of feat] (tab) {Feature Matrix\\features.csv};
\node[io, fill=lime!12, draw=lime!45!black, below=of det] (alert) {Outputs\\metrics, plots, model artifacts};

\draw[arr] (cfg) -- (gen);
\draw[arr] (gen) -- (val);
\draw[arr] (val) -- (feat);
\draw[arr] (feat) -- (det);
\draw[arr] (val) |- (dset);
\draw[arr] (feat) -- (tab);
\draw[arr] (det) -- (alert);
\end{tikzpicture}
}
\caption{AegisUI pipeline. Each stage reads the prior output and produces its own artifacts.}
\label{fig:architecture}
\end{figure*}
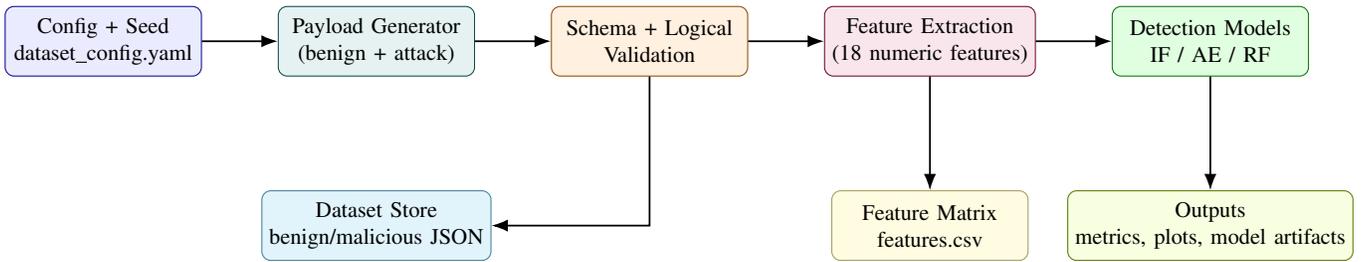

\textbf{Generation.} Benign payloads start from domain blueprints. For each domain (booking assistant, e-commerce, analytics dashboard, form submission, workflow approval), the generator creates a root container and grows the tree stochastically within configured bounds: 5--40 components, depth 1--5. Component types are drawn from a fixed vocabulary of ten (Button, TextField, Card, Table, Form, Modal, Dropdown, Checkbox, Label, Chart). Each component gets a role (container, action, input, or display) with domain-appropriate labels and properties. Bindings connect input and display components to domain-scoped data sources like \texttt{booking\_assistant.flight\_ref}.

Malicious payloads are never generated from scratch. We pick a benign payload and mutate it using one of five attack strategies. This matters because malicious payloads share the same base distribution as benign ones, which makes detection harder and more realistic than if we had produced attack samples independently.

\textbf{Validation.} Every payload goes through schema validation (nine required top-level fields, consistent component IDs, correct types) and logical validation (depth within bounds, table metadata present, form-input relationships intact). Invalid payloads are logged and discarded. In practice, under 0.5\% of generated payloads failed validation.

\textbf{Feature extraction.} Each valid payload becomes an 18-dimensional numeric vector. See Section~\ref{sec:features}.

\textbf{Detection.} Three models train and evaluate on the same stratified split.

\section{Dataset}
\subsection{Composition}
We generated 4{,}000 payloads: 3{,}000 benign and 1{,}000 malicious. The 3:1 ratio is deliberate because in most agent systems the vast majority of payloads are legitimate, so a balanced dataset would not reflect reality.

Table~\ref{tab:dataset-stats} gives the summary. Average component count is 19.77, ranging from 4 (a minimal form) to 85 (a deeply nested dashboard with attack-injected nodes). Average depth is 2.64, with tails reaching~10 on layout-abuse payloads.

\begin{table}[t]
\caption{Dataset overview}
\label{tab:dataset-stats}
\centering
\begin{tabular}{l r}
\toprule
\textbf{Statistic} & \textbf{Value} \\
\midrule
Total samples & 4{,}000 \\
Benign & 3{,}000 \\
Malicious & 1{,}000 \\
Domains & 5 \\
Attack families & 5 \\
Seed & 1337 \\
Mean component count & 19.77 \\
Mean max depth & 2.64 \\
Mean bindings per payload & 9.43 \\
\bottomrule
\end{tabular}
\end{table}

Domains are roughly balanced: analytics dashboard (832), form submission (830), e-commerce (789), booking assistant (777), workflow approval (772).

\subsection{Attack Families}
Table~\ref{tab:attack-dist} summarizes the five attack types and what each one does in practice.

\begin{table}[t]
\caption{Attack type distribution}
\label{tab:attack-dist}
\centering
\begin{tabular}{l r p{4.0cm}}
\toprule
\textbf{Attack Type} & \textbf{Count} & \textbf{Mechanism} \\
\midrule
workflow\_anomaly & 258 & Reorders actions so approvals fire before input validation; marks components ``approved without review'' \\
phishing\_interface & 232 & Injects credential/payment fields (email, password, credit card, fake ``secure validation'' button) into a benign container \\
data\_leakage & 228 & Rebinds display widgets to internal sources like \texttt{internal.user\_ssn} or \texttt{internal.auth\_secret} \\
manipulative\_ui & 207 & Swaps button labels to benign text while the hidden action is destructive (\texttt{delete\_account}) \\
layout\_abuse & 75 & Nests 6--12 extra container layers and floods them with 15--40 filler components \\
\bottomrule
\end{tabular}
\end{table}

Layout abuse is underrepresented (75 samples) because it produces the most aggressive structural changes and the generator's validation gate catches some extreme cases. We kept the natural distribution rather than forcing balance, since that better reflects how attack frequencies might vary in practice.

Each malicious payload records full provenance: the benign payload it was mutated from, the attack type, a severity label, and a trace listing exactly which components were injected or modified. When a model gets a prediction wrong, we can look at the trace and understand exactly what happened.

\subsection{Why Synthetic Data}
No public dataset captures structured, agent-generated UI protocol payloads with labeled attack semantics. Collecting one from production would require access to live agent systems and user interaction logs with manual annotation, none of which is openly available.

Synthetic generation gives us exact ground truth labels, full reproducibility through seed control, and controlled attack coverage. The trade-off is ecological validity: real systems have broader component vocabularies, multilingual labels, longer session histories, and adversaries who iterate. We are explicit about this throughout.

Sommer and Paxson \cite{somer2010outside} warned about ML security work that overstates synthetic results. We tried to take that seriously. Our numbers are a controlled baseline, not a deployment guarantee.

\section{Feature Engineering}
\label{sec:features}
We extract 18 numeric features from each payload, organized into four groups. Table~\ref{tab:features} lists them all.

\begin{table}[t]
\caption{All 18 features, grouped by category}
\label{tab:features}
\centering
\footnotesize
\begin{tabular}{l l}
\toprule
\textbf{Feature} & \textbf{Group} \\
\midrule
component\_count & Structural \\
unique\_component\_types & Structural \\
max\_depth & Structural \\
avg\_branching\_factor & Structural \\
graph\_density & Structural \\
payload\_size\_bytes & Structural \\
container\_ratio & Structural \\
action\_component\_ratio & Structural \\
\midrule
avg\_label\_length & Semantic \\
text\_entropy & Semantic \\
sensitive\_keyword\_count & Semantic \\
semantic\_inconsistency\_score & Semantic \\
numeric\_property\_statistics & Semantic \\
\midrule
number\_of\_bindings & Binding \\
sensitive\_binding\_flag & Binding \\
cross\_component\_binding\_ratio & Binding \\
\midrule
timestamp\_variance & Session \\
inter\_payload\_interval & Session \\
\bottomrule
\end{tabular}
\end{table}

\textbf{Structural (8 features).} These describe the shape of the payload's component tree. Component count and payload size in bytes are the most obvious. Max depth measures nesting, and layout-abuse attacks push this well above normal levels. Graph density and branching factor capture how spread out the tree is. Container ratio and action-component ratio give the proportion of components serving those roles.

\textbf{Semantic (5 features).} Average label length and Shannon entropy of label text describe the wording. Sensitive keyword count scans labels and bindings for words like ``password'', ``credit card'', ``ssn'', ``token'', and ``secret.'' The semantic inconsistency score is probably our most targeted feature: for every action component, we check whether a benign-sounding label (``View invoice'', ``Continue safely'') is paired with a risky hidden action (\texttt{delete\_account}, \texttt{authorize\_transfer}). The score is the count of such mismatches divided by total component count. Numeric property statistics averages the numeric values found in component properties (table row counts, max-length constraints, etc.).

\textbf{Binding (3 features).} Number of bindings is straightforward. The sensitive binding flag is binary and fires when any binding target contains keywords like ``ssn'', ``payment'', or ``secret.'' Cross-component binding ratio measures how often multiple components share the same data source.

\textbf{Session (2 features).} Timestamp variance and inter-payload interval capture session-level timing. These turned out to be the weakest predictors. We think the reason is simple: synthetic sessions are short (typically 1--3 payloads), so there is barely any temporal signal to exploit. With longer real-world sessions, they might carry more weight.

\section{Detection Models}
We chose three models to span different label-availability assumptions.

\subsection{Isolation Forest}
Isolation Forest \cite{liu2008isolation} partitions the feature space randomly. Points in sparse regions get isolated quickly, so a short average path implies a likely anomaly. We used 300 trees with scikit-learn's \cite{pedregosa2011sklearn} auto contamination setting. No labels needed.

The downside showed up quickly in our experiments: Isolation Forest mainly catches geometric outliers. If an attack payload's features sit inside the benign distribution's convex hull, it goes undetected. Manipulative-UI payloads, which modify only one or two button labels, are exactly that kind of subtle.

\subsection{Autoencoder}
The autoencoder learns what benign looks like. We train it on only the benign portion of the training set. At test time, high reconstruction error $\to$ anomaly.

Architecture: input dim = 18 (one per feature). Encoder: Linear(18$\to$16) $\to$ ReLU $\to$ Linear(16$\to$8) $\to$ ReLU. Decoder: Linear(8$\to$16) $\to$ ReLU $\to$ Linear(16$\to$18). Bottleneck is 8 units. We trained for 80 epochs, batch size 64, Adam optimizer with learning rate $10^{-3}$, MSE loss. The decision threshold was the 95th percentile of benign training-set reconstruction errors (threshold value: 0.0874).

We tried bottleneck sizes of 4, 8, and 16 during development. Four units was too aggressive because benign reconstruction error shot up so much that the threshold became useless (too many false positives). Sixteen units was too generous, since attack payloads also reconstructed well, making them hard to separate. Eight was the sweet spot.

\subsection{Random Forest}
Random Forest \cite{breiman2001random}: 400 trees, class-weight balancing (to handle the 3:1 imbalance), no max-depth constraint, scikit-learn implementation. It sees labels during training.

This is the ceiling. If you have labeled attack data, a supervised ensemble will beat unsupervised methods. We include it not for novelty but to set the upper bound and to get feature importances, which tell us what actually drives the predictions.

\section{Experimental Setup}
We split the 4{,}000-sample feature matrix 80/20 using stratified sampling (preserving the 3:1 class ratio), giving 3{,}200 training and 800 test samples. All three models evaluated on the same test set.

Feature normalization: z-score scaling (zero mean, unit variance) fitted on training data and applied to test data. This matters most for the autoencoder, because MSE loss is sensitive to feature scale.

The autoencoder trained on 2{,}400 benign training samples only. Isolation Forest and Random Forest used the full 3{,}200-sample training set.

All random operations used seed 1337. Implementation: Python 3.12, scikit-learn \cite{pedregosa2011sklearn}, PyTorch \cite{paszke2019pytorch}, Pandas, NumPy.

Metrics: accuracy, precision, recall, F1-score, ROC-AUC, and confusion matrices. We report confusion matrices explicitly because aggregate numbers hide failure modes.

\section{Results}
\label{sec:results}
\subsection{Overall Performance}
Table~\ref{tab:model-results} has the numbers. Random Forest leads across the board.

\begin{table}[t]
\caption{Model performance on 800-sample test set}
\label{tab:model-results}
\centering
\begin{tabular}{lccccc}
\toprule
\textbf{Model} & \textbf{Acc.} & \textbf{Prec.} & \textbf{Rec.} & \textbf{F1} & \textbf{AUC} \\
\midrule
Isolation Forest & 0.824 & 0.757 & 0.435 & 0.552 & 0.822 \\
Autoencoder & 0.885 & 0.790 & 0.735 & 0.762 & 0.863 \\
Random Forest & \textbf{0.931} & \textbf{0.980} & \textbf{0.740} & \textbf{0.843} & \textbf{0.952} \\
\bottomrule
\end{tabular}
\end{table}

The confusion matrices tell a sharper story. Random Forest produced only 3 false positives out of 600 benign test samples (0.5\% FP rate) while catching 148 of 200 malicious ones. That is a very conservative boundary; it almost never cries wolf, which matters in production where alert fatigue kills adoption.

The autoencoder caught nearly the same number of attacks (147 vs.\ 148) but at the cost of 39 false positives, which is about 13$\times$ more false alarms. Still, a 6.5\% FP rate is manageable during a monitoring-only rollout.

Isolation Forest was the weakest. It flagged 87 of 200 malicious samples and let 113 through, missing over half of them. Decent AUC (0.822) was masked by poor recall.

\begin{figure}[t]
\centering
\subfloat[Isolation Forest]{\includegraphics[width=0.31\linewidth]{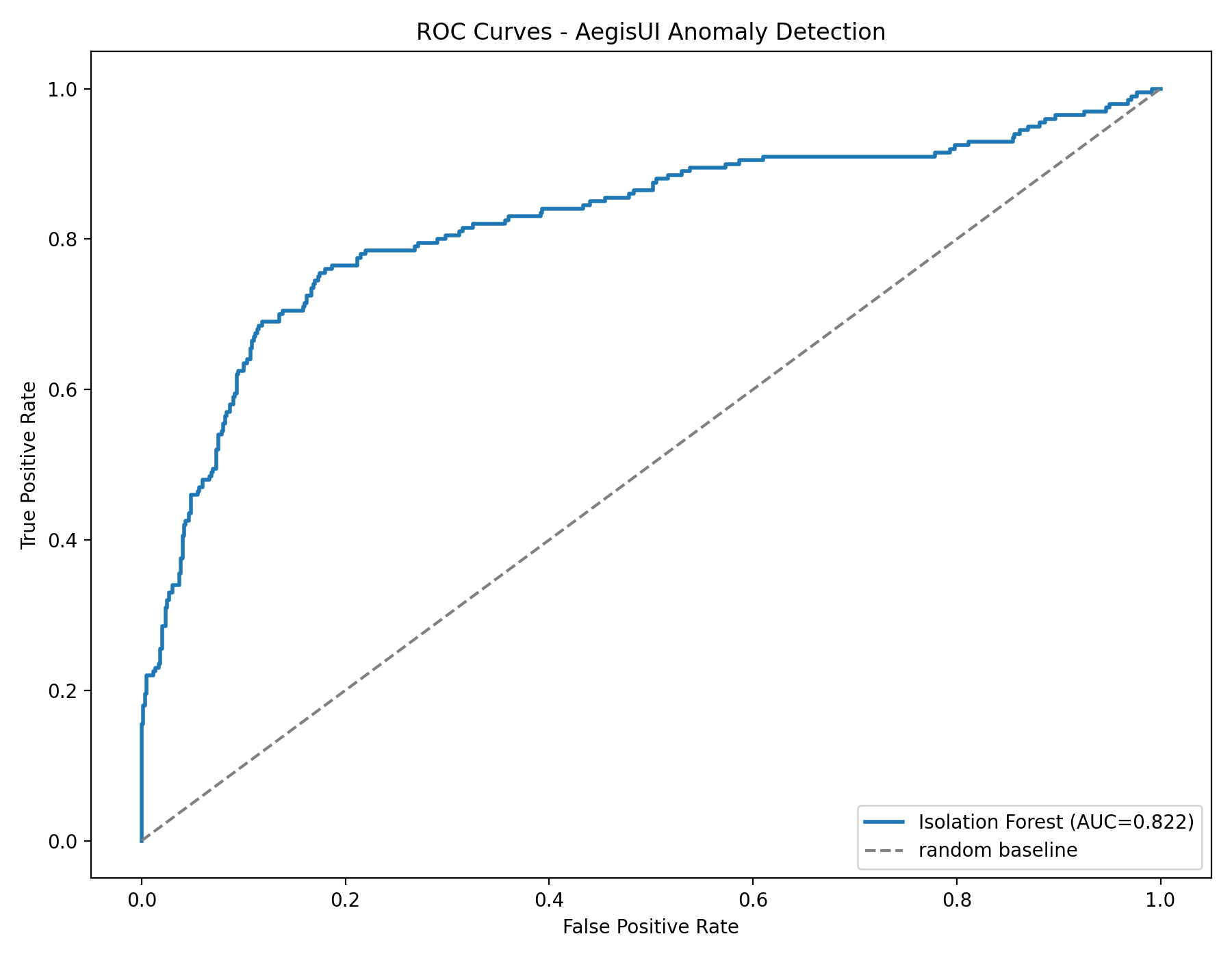}}
\hfill
\subfloat[Autoencoder]{\includegraphics[width=0.31\linewidth]{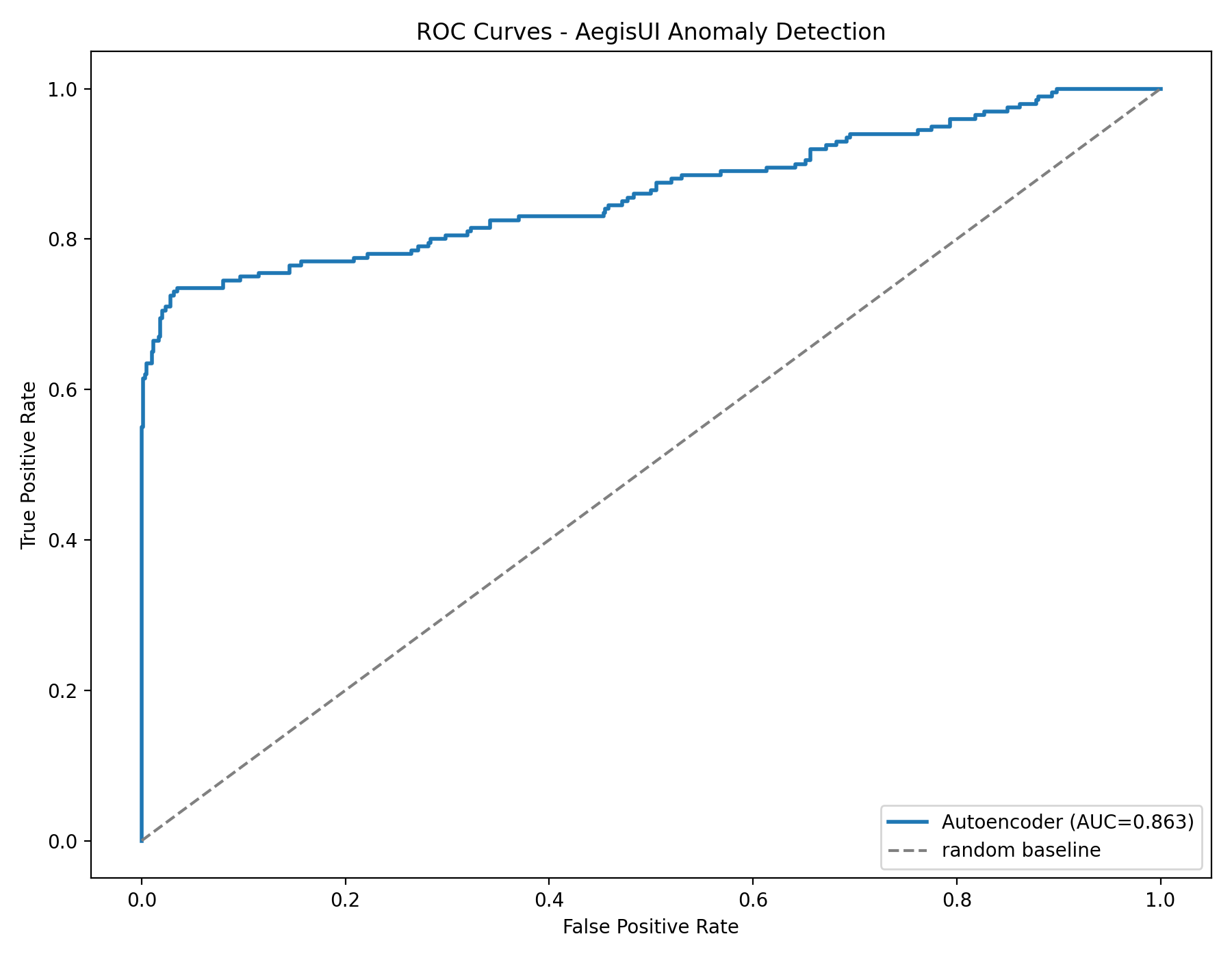}}
\hfill
\subfloat[Random Forest]{\includegraphics[width=0.31\linewidth]{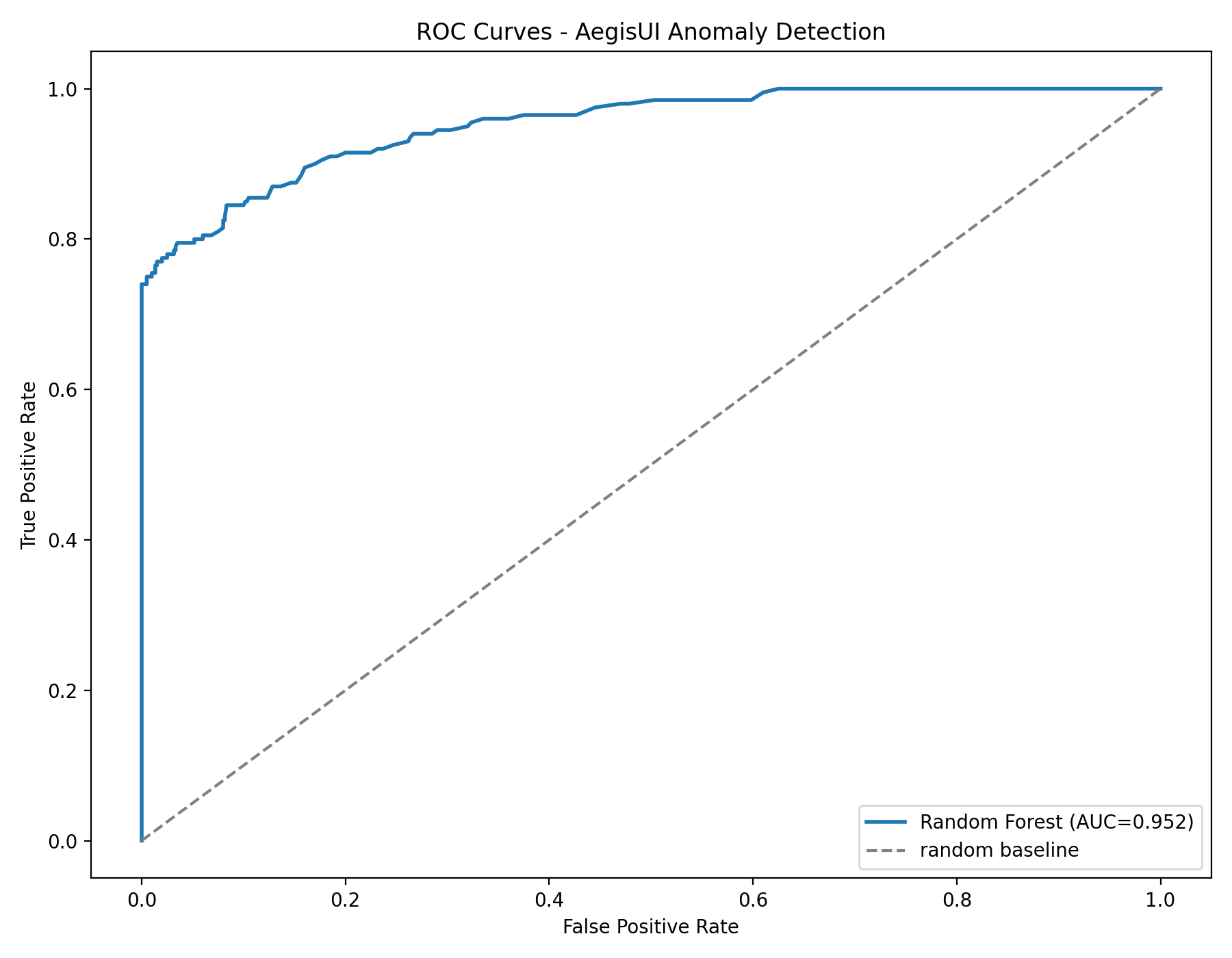}}
\caption{ROC curves for the three models.}
\label{fig:roc-compare}
\end{figure}

\begin{figure}[t]
\centering
\includegraphics[width=0.98\linewidth]{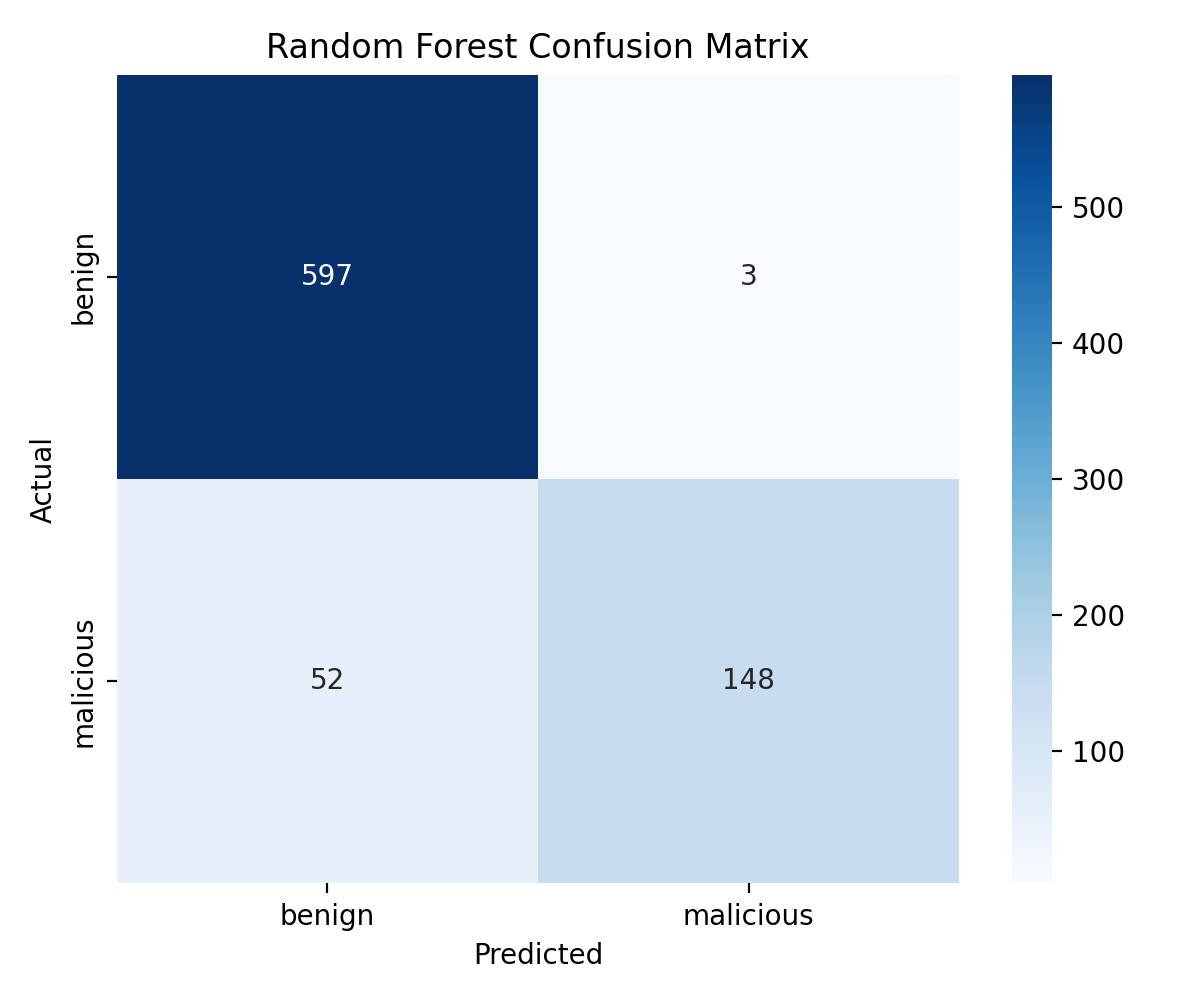}
\caption{Random Forest confusion matrix. TN=597, FP=3, FN=52, TP=148.}
\label{fig:cm-rf}
\end{figure}

\subsection{Per-Attack-Type Breakdown}
Not all attacks are equally hard. We examined which attack families Random Forest caught and which it missed.

\textbf{Layout abuse:} easiest to detect. Every layout-abuse sample in the test set was correctly flagged. These payloads have 50+ components, depth above 6, and abnormal graph density. Structural features alone handle them.

\textbf{Phishing interface:} detected reliably. The sensitive keyword count and sensitive binding flag fire strongly because the injected components contain terms like ``password'' and ``credit card.''

\textbf{Data leakage:} solid but not perfect. When the leakage involves keywords (\texttt{internal.auth\_secret}), the binding features catch it. When the target is a subtler internal field name without flagged keywords, the detector misses it.

\textbf{Workflow anomaly:} mid-range. The model picks up reordered action properties and ``approved without review'' markers, but the signal is weaker since these attacks modify existing components rather than injecting new ones.

\textbf{Manipulative UI:} hardest. These payloads change just a few button labels and their hidden action properties. The structural footprint is nearly identical to benign. The semantic inconsistency score catches some cases, but when only one or two buttons are tampered in a 20-component payload, the normalized score is tiny.

\subsection{Feature Importance}
\begin{figure}[t]
\centering
\includegraphics[width=0.98\linewidth]{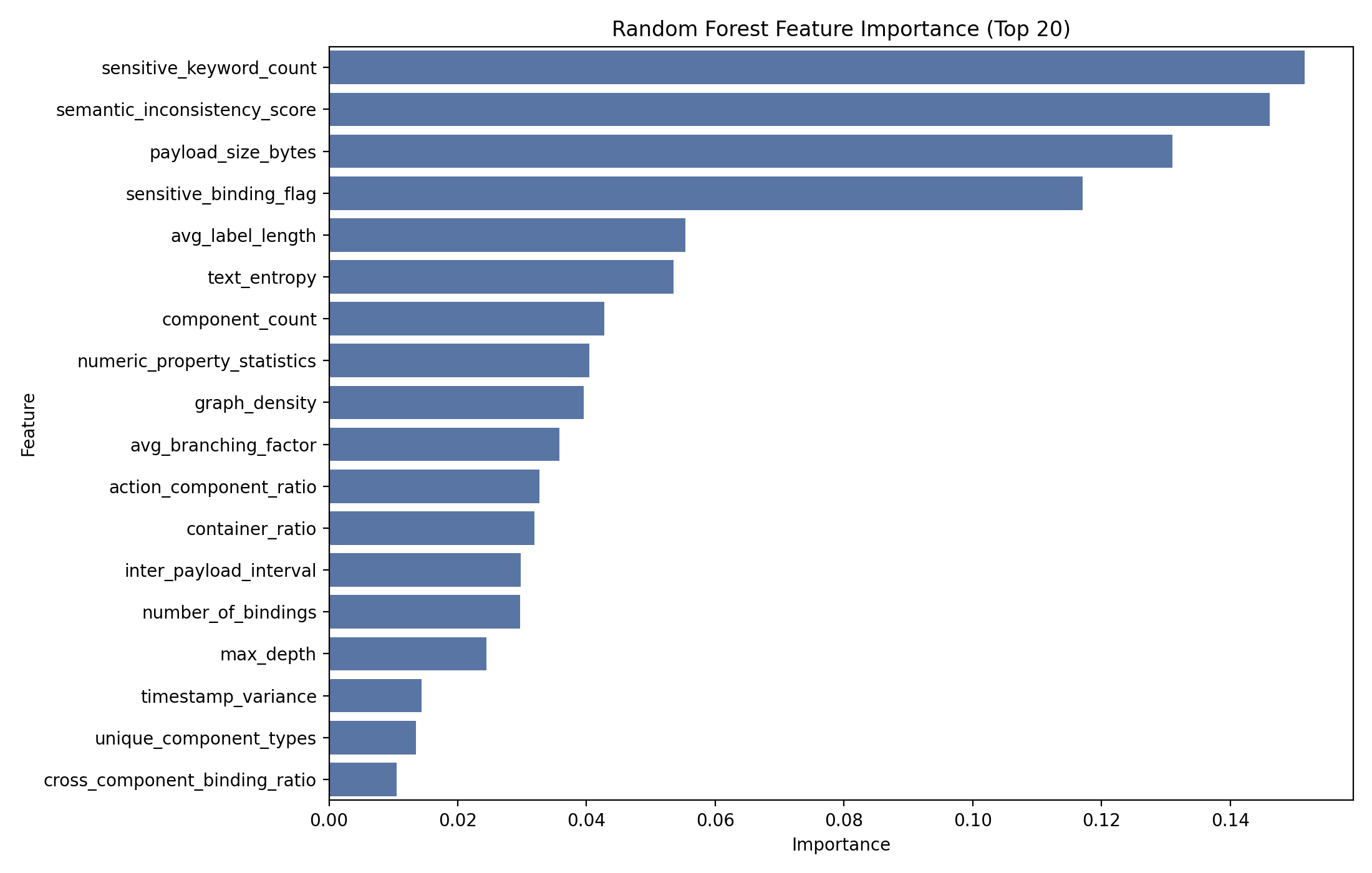}
\caption{Random Forest feature importances.}
\label{fig:fi-rf}
\end{figure}

Fig.~\ref{fig:fi-rf} shows Random Forest feature importances.

Structural features dominate: component count, payload size, max depth, graph density. This partly reflects the strong signal from layout-abuse attacks, but even without those, phishing injection also adds components and changes tree shape.

Sensitive keyword count and the sensitive binding flag rank mid-tier: strong for phishing and leakage, irrelevant for layout abuse and workflow anomalies.

Session features (timestamp variance, inter-payload interval) ranked last. We expected timing patterns to help and were surprised when they did not. The explanation is probably that our synthetic sessions are too short for temporal patterns to develop.

\subsection{Feature-Group Ablation}
Table~\ref{tab:ablation} shows Random Forest retrained on each feature group alone.

\begin{table}[t]
\caption{Ablation: Random Forest with feature subsets}
\label{tab:ablation}
\centering
\begin{tabular}{lcc}
\toprule
\textbf{Feature Subset} & \textbf{F1} & \textbf{AUC} \\
\midrule
Structural only (8 features) & 0.772 & 0.904 \\
Semantic only (5 features) & 0.691 & 0.852 \\
Binding only (3 features) & 0.642 & 0.811 \\
All 18 features & \textbf{0.843} & \textbf{0.952} \\
\bottomrule
\end{tabular}
\end{table}

Structural features alone get 0.772 F1. Adding semantic and binding features bumps it to 0.843, a 9\% improvement. Each group contributes, but structural does the heavy lifting. Binding features on their own are weakest (0.642 F1), which we think is because there are only three of them and the binary sensitive binding flag lacks resolution.

\section{Discussion}
\textbf{The autoencoder is the practical story.} Random Forest wins the numbers, but that requires labeled attack data. The autoencoder achieves 0.762 F1 without ever seeing a malicious payload during training. If you are rolling out a new agent system and have zero attack history, you can train the autoencoder on your benign traffic from day one and have a usable detector. You can always switch to a supervised model later when you have accumulated enough labeled incidents. We think this is the more relevant finding for real deployment.

\textbf{The 52 missed attacks are not random.} Random Forest's false negatives cluster in manipulative UI and some data leakage samples, namely attacks that change one or two components in an otherwise normal payload. Our tabular features are computed over the entire payload and averaged or normalized by component count, which dilutes local signals. A model that operates on individual components or subgraphs would have a better shot at catching these.

\textbf{False positive cost.} The 3 false positives from Random Forest (0.5\% FP rate) are likely acceptable in production. The autoencoder's 39 (6.5\%) would be fine during a logging-only phase but would cause friction in a hard-block deployment. The choice depends on where the system sits on the ``block aggressively'' vs.\ ``monitor and alert'' spectrum.

\textbf{Limitations we want to be honest about.} The dataset is synthetic, covering five domains, five attack types, and ten component types. Real agents will have richer vocabularies, domain-specific jargon, and adversaries who specifically try to evade detection. We have not tested adversarial evasion; an attacker who knows our feature set could flatten their structural footprint and avoid keyword triggers. The single train-test split, while stratified, does not give confidence intervals; $k$-fold cross-validation would strengthen the results and is planned for a follow-up.

\section{Conclusion}
AegisUI shows that behavioral anomaly detection on structured UI protocol payloads is feasible with relatively simple features and off-the-shelf models. On 4{,}000 payloads, Random Forest achieves 0.843 F1 and 0.952 AUC. The autoencoder, needing no attack labels, gets 0.762 F1, which is a viable starting point for systems without labeled attack history.

The harder finding is where detection breaks down. Attacks that modify only a small part of a large payload produce weak aggregate signals. All three models struggle with them, which points at the next step: component-level or graph-level representations instead of payload-level tabular features.

We release all code, data, configuration, and trained models so others can reproduce, extend, and compare. The pipeline is deterministic: same seed, same output.

Three directions we plan to pursue next:
\begin{itemize}
    \item \textbf{Graph neural networks.} Specifically GraphSAGE \cite{graphsage} on the layout graph, with per-node features from individual component properties. This should catch localized anomalies, such as a single manipulated button, that get washed out in aggregate features.
    \item \textbf{Session sequence modeling.} Our two session features are too simple. With longer sessions from real systems, an LSTM or Transformer over payload sequences could detect statistical departures from an agent's typical interaction pattern.
    \item \textbf{Mixed synthetic-real evaluation.} We plan to validate synthetic-trained models on anonymized protocol traces from operational agent systems to measure the domain-adaptation gap.
\end{itemize}

\begin{figure}[t]
\centering
\begin{tikzpicture}[
node distance=0.8cm and 0.55cm,
blk/.style={draw, rounded corners, align=center, minimum width=0.95\linewidth, minimum height=0.72cm},
arr/.style={-{Latex[length=2.2mm]}, thick}
]
\node[blk, fill=purple!10, draw=purple!65!black] (g) {UI Layout Graph\\(nodes = components, edges = parent-child)};
\node[blk, fill=blue!10, draw=blue!55!black, below=of g] (enc) {GraphSAGE Encoder\\(node-level message passing)};
\node[blk, fill=teal!12, draw=teal!55!black, below=of enc] (temp) {Session Sequence Model\\(LSTM over payload embeddings)};
\node[blk, fill=green!12, draw=green!45!black, below=of temp] (head) {Risk Scorer + Policy Enforcement};
\draw[arr] (g) -- (enc);
\draw[arr] (enc) -- (temp);
\draw[arr] (temp) -- (head);
\end{tikzpicture}
\caption{Planned next-generation architecture: graph + sequence modeling.}
\label{fig:future-gnn}
\end{figure}
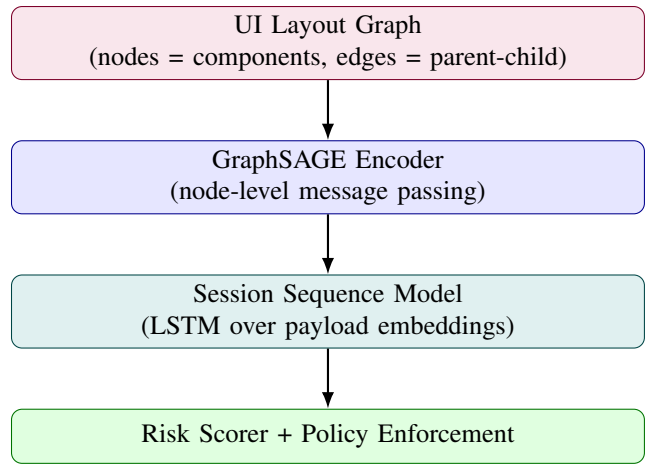

\end{document}